\documentclass[letterpaper, 10 pt, conference]{ieeeconf}  % Comment this line out if you need a4paper

\IEEEoverridecommandlockouts                              % This command is only needed if 
\usepackage[linesnumbered,ruled,vlined]{algorithm2e}
\usepackage{algorithmic}
\usepackage{graphics} % for pdf, bitmapped graphics files
\graphicspath{{figures/}}
\usepackage{tikz}
\usepackage{graphicx}
\usepackage{comment}
\usepackage{epsfig} % for postscript graphics files
\usepackage{times} % assumes new font selection scheme installed
\usepackage{siunitx}
\usepackage{amsmath} % assumes amsmath package installed
\usepackage{amssymb}  % assumes amsmath package installed
\usepackage{hyperref}
\usepackage[capitalise]{cleveref}
\Crefname{algorithm}{Alg.}{Algs.}
\Crefname{section}{Sec.}{Secs.}
\Crefname{equation}{Eq.}{Eqs.}
\usepackage{authblk}
\usepackage{booktabs}
\usepackage[
    sorting=none,
    giveninits=true,
    maxbibnames=9,
    maxcitenames=2,backend=biber
    ]{biblatex}
\renewcommand{\baselinestretch}{0.98}

\title{\LARGE \bf
LaCE-LHMP: Airflow Modelling-Inspired Long-Term Human Motion Prediction By Enhancing Laminar Characteristics in Human Flow
}

\author{
Yufei Zhu$^{1}$, Han Fan$^{1}$, Andrey Rudenko$^{2}$, Martin Magnusson$^{1}$, Erik Schaffernicht$^{1}$, Achim J. Lilienthal$^{1,3}$%
\thanks{$^{1}$Robot Navigation and Perception Lab, AASS Research Center, 
{\"O}rebro University, Sweden {\tt\small yufei.zhu@oru.se; han.fan@oru.se}}%
\thanks{$^{2}$Robert Bosch GmbH, Corporate Research, Stuttgart, Germany}%
\thanks{$^{3}$Chair: Perception for Intelligent Systems, Technical University of Munich, Germany}%
\thanks{This work has received funding from the European Union’s Horizon 2020 research and innovation programme under grant agreement No 101017274 (DARKO), and is also partially funded by the academic program Sustainable Underground Mining (SUM) project, jointly financed by LKAB and the Swedish Energy Agency.}
}

\begin{document}

\maketitle
\thispagestyle{empty}
\pagestyle{empty}

%%%%%%%%%%%%%%%%%%%%%%%%%%%%%%%%%%%%%%%%%%%%%%%%%%%%%%%%%%%%%%%%%%%%%%%%%%%%%%%%
\begin{abstract}
Long-term human motion prediction (LHMP) is essential for safely operating autonomous robots and vehicles in populated environments. It is fundamental for various applications, including motion planning, tracking, human-robot interaction and safety monitoring. However, accurate prediction of human trajectories is challenging due to complex factors, including, for example, social norms and environmental conditions. The influence of such factors can be captured through Maps of Dynamics (MoDs), which encode spatial motion patterns learned from (possibly scattered and partial) past observations of motion in the environment and which can be used for data-efficient, interpretable motion prediction (MoD-LHMP). To address the limitations of prior work, especially regarding accuracy and sensitivity to anomalies in long-term prediction, we propose the Laminar Component Enhanced LHMP approach (LaCE-LHMP). Our approach is inspired by data-driven airflow modelling, which estimates laminar and turbulent flow components and uses predominantly the laminar components to make flow predictions. Based on the hypothesis that human trajectory patterns also manifest laminar flow (that represents predictable motion) and turbulent flow components (that reflect more unpredictable and arbitrary motion), LaCE-LHMP extracts the laminar patterns in human dynamics and uses them for human motion prediction. We demonstrate the superior prediction performance of LaCE-LHMP through benchmark comparisons with state-of-the-art LHMP methods, offering an unconventional perspective and a more intuitive understanding of human movement patterns.
\end{abstract}

\section{INTRODUCTION}

\begin{figure}
\centering
\includegraphics[width=.42\linewidth]{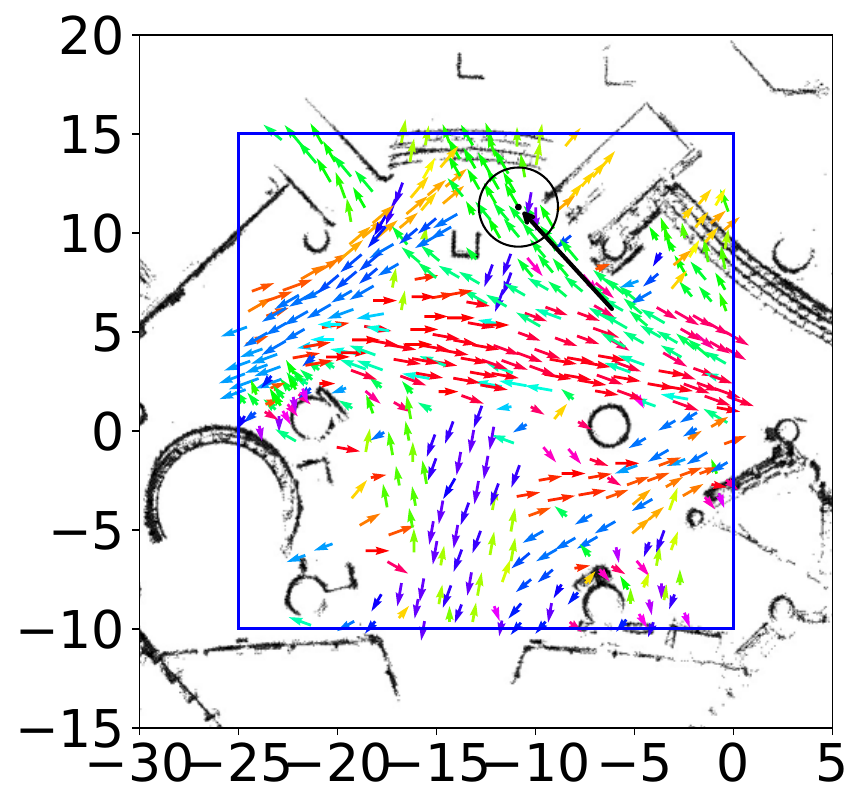}
\hspace{3.4mm}
\includegraphics[width=.465\linewidth]{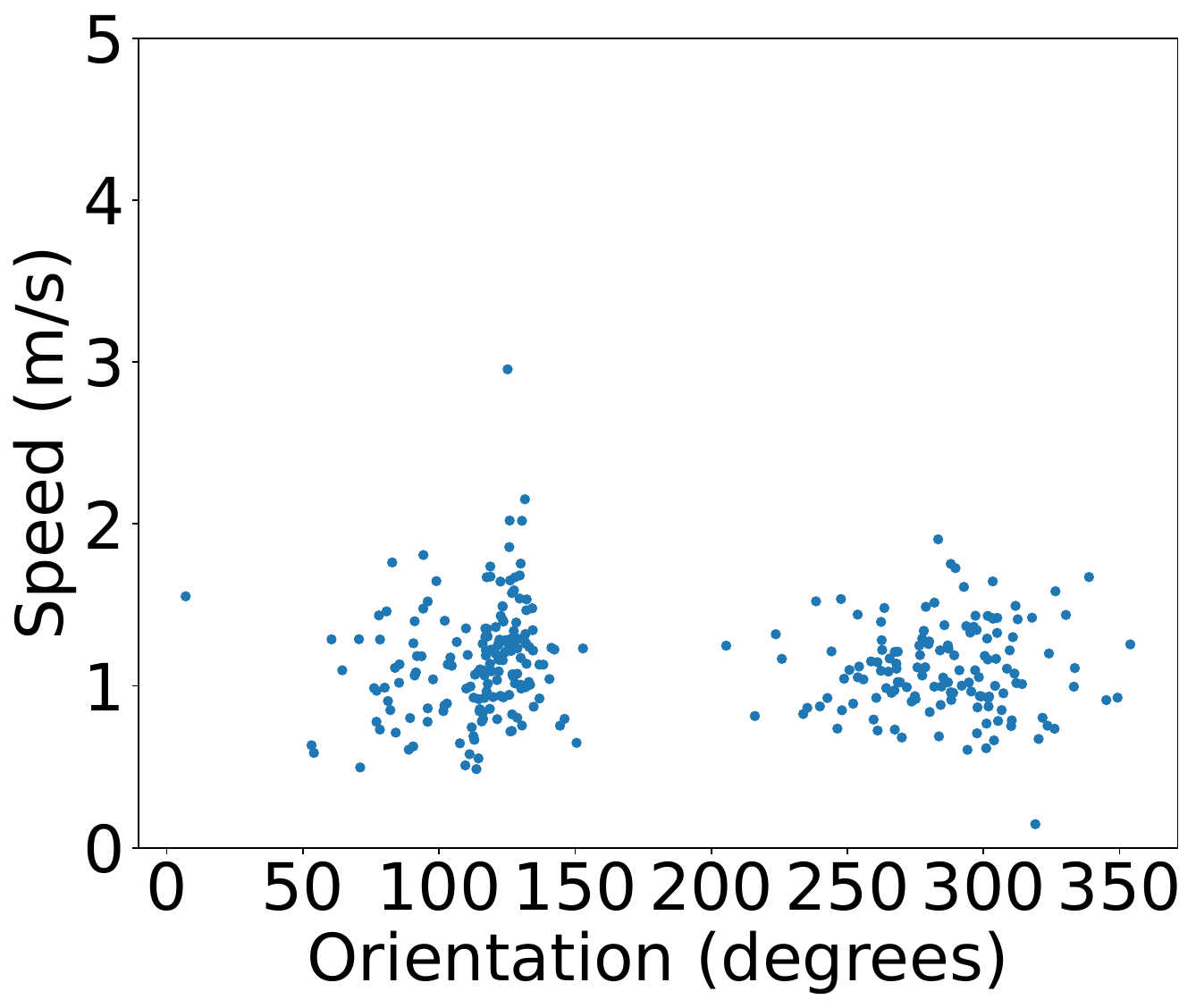}
\includegraphics[width=.45\linewidth]{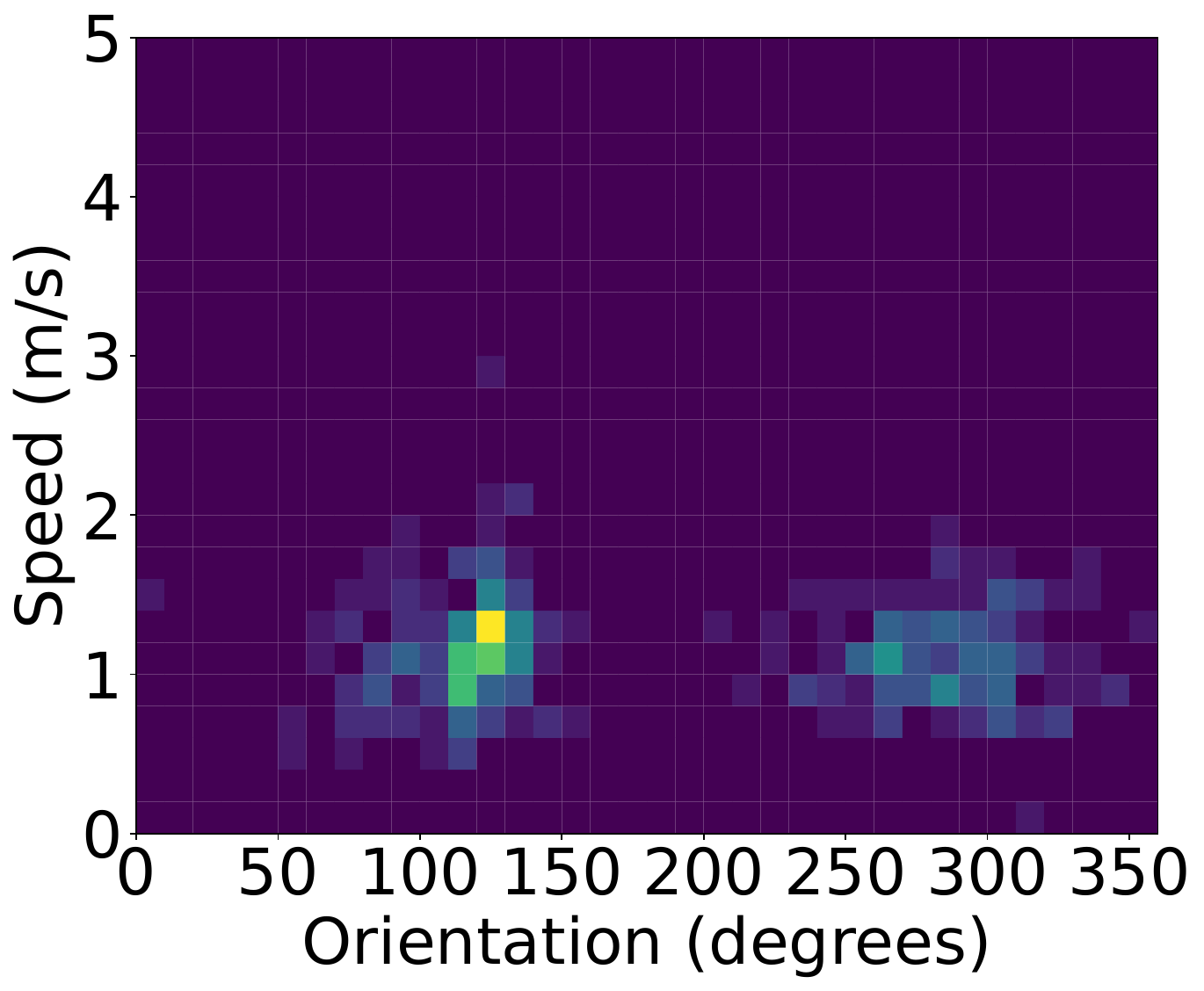}
\hspace{2.5mm}
\includegraphics[width=.45\linewidth]{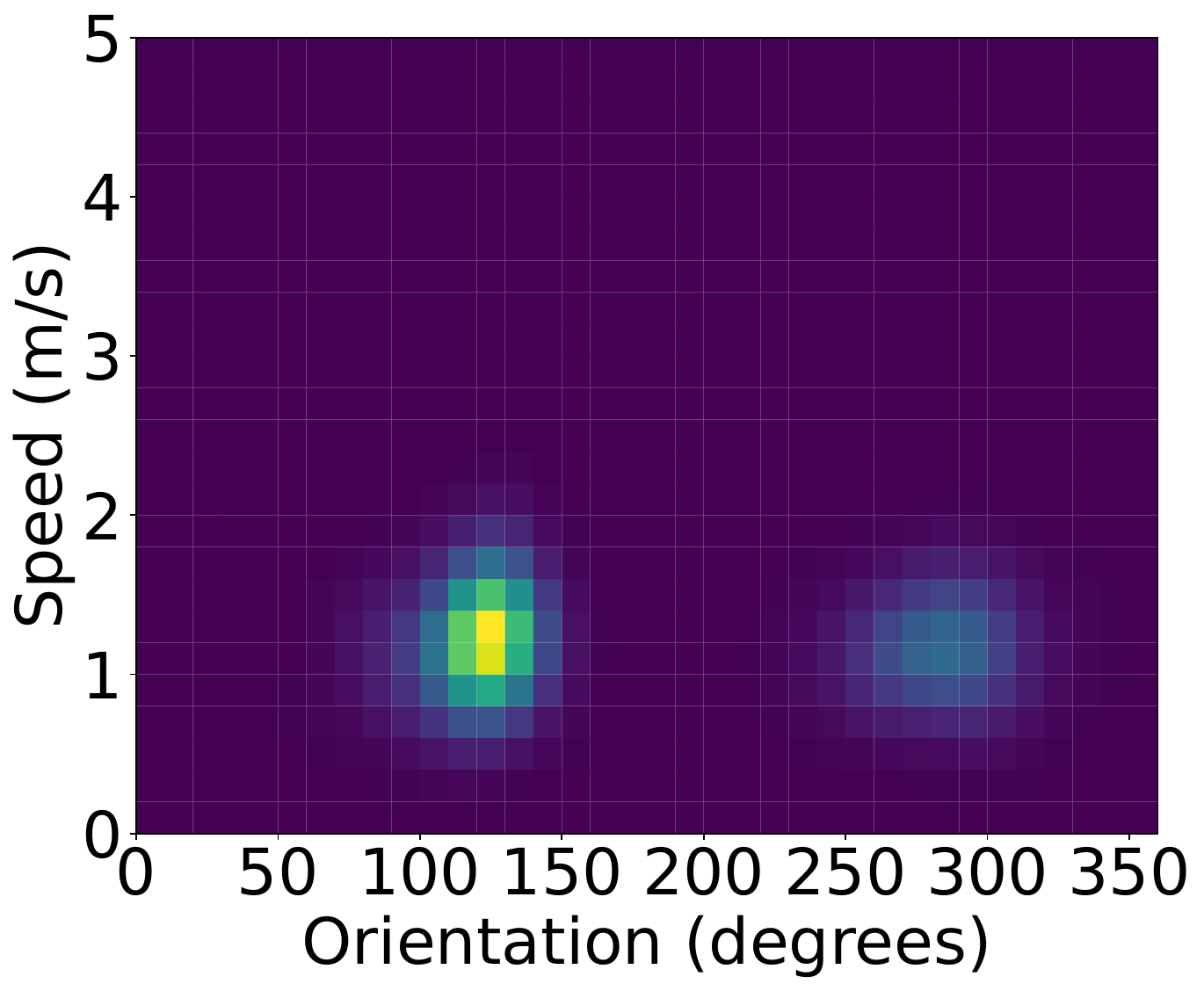}

\caption{Example of laminar component extraction in LaCE-LHMP. \textbf{Upper-left}: LaCE model of a location in a shopping mall. Colored arrows show flow directions with highest likelihoods; \textbf{Upper-right}: raw data (velocity observations) in the $\omega-\nu$ domain (i.e. speed and orientation) at a specific location; \textbf{Lower-left:} histogram of the raw data $\Gamma^R$; \textbf{Lower-right}: extracted laminar component $\Gamma^L$. The laminar component is used for motion prediction in LaCE-LHMP.}
\label{fig:laminar_extraction_example}
\vspace*{-4mm}
\end{figure}

Long-term human motion prediction (LHMP) plays an important role in ensuring the safe operation of autonomous robots and vehicles in populated environments~\cite{rudenko2020human}. Accurate prediction of people's future trajectories over a prolonged duration stands as a fundamental requirement for various applications, including optimized motion planning, refined tracking, advanced automated driving, improved human-robot interaction, and enhanced intelligent safety monitoring and surveillance. Accurate LHMP not only improves operational efficiency in relevant applications but also fosters a higher level of acceptance among users and stakeholders, as they can trust the systems to understand and anticipate human motion more reliably.

Human motion is complex, influenced by various factors, including not only an individual's intrinsic intent and dynamics but also external influences such as social conventions and environmental cues. These factors collectively contribute to the challenge of accurately predicting human motion~\cite{rudenko2020human}. Especially for predictions over an extended, very long time horizon (20 seconds and more), the impact of complex, large-scale environments on human behavior needs to be accounted for. Unlike for short-term predictions, where considering only the current state and immediate interactions can suffice, the long-term perspective demands explicit modelling of how the environment continuously shapes and directs human motion. These influences cannot be adequately summarized just by the current state of the individual and observed interactions but instead require explicit modelling~\cite{rudenko2018human}. 

An effective approach to address this challenge is to use \emph{maps of dynamics} (MoDs). MoDs \cite{kucner2023survey} are maps that encode spatial or spatio-temporal motion patterns as a feature of the environment. MoD-informed long-term human motion prediction (MoD-LHMP) approaches are particularly suited to predict motion in the long-term perspective, where the environment effects become critical for making accurate predictions. By using MoDs, motion prediction can utilize previously observed spatial motion patterns that encode important information about spatial motion patterns in a given environment. Among the MoD-LHMP methods, Zhu~et~al.~\cite{zhu2023clifflhmp} utilize CLiFF-maps \cite{kucner2017enabling}, which capture multimodal statistical information about human flow patterns, to make long term predictions. CLiFF-LHMP is shown to make accurate long-term predictions, even when trained with small amounts of data \cite{zhu2023data}. However, the modelling approach based on the CLiFF map may struggle to differentiate dominant human flow from irregular motion, and therefore the prediction accuracy may be affected by anomalous data. 

Detecting and identifying abnormal trajectories is a major challenge in motion modelling and prediction. Existing methods typically identify abnormal motions by comparing them to expected behaviors \cite{liu2018ano_pred} or measuring deviations from normal motions \cite{fernando2018soft}. However, these approaches require labelled data for supervised learning.

In this paper, we propose the Laminar Component Enhanced (LaCE) LHMP approach inspired by data-driven airflow modelling \cite{bennetts2017probabilistic}. Airflow can be characterized as a combination of laminar and turbulent flow patterns in fluid dynamics~\cite{ashrae2017fundamentals,doolan2022laminar}. Similarly, we postulate that human trajectory patterns share this property, with laminar components representing predictable motion and turbulent flow components reflecting more unpredictable and arbitrary motion. Accordingly, the proposed LaCE-LHMP approach extracts laminar patterns in human dynamics and uses them for motion prediction, mitigating the impact of anomalous data in an unsupervised manner. During the prediction process, the degree of laminar dominance is quantitatively measured to make adaptive adjustments to the contribution of the laminar component. LaCE-LHMP also ranks the predicted trajectories and provides the most likely output, offering practical utility for autonomous robots. Our approach shares the benefits of the prior art in MoD-LHMP, while addressing its limitations.

%By focusing on the laminar component of human dynamics, which represents more streamlined and predictable flows, we can distill the dominant movement trajectories, akin to the primary currents in a water body. This approach offers not only a novel perspective but promises enhanced accuracy and a more intuitive understanding of human movement patterns.

We demonstrate the proposed approach in quantitative and qualitative experiments, comparing it to several state-of-the-art LHMP methods. The superior prediction accuracy is promising and supports the hypothesis that human motion in real-world environments comprises underlying laminar patterns.
We note that the LaCE-LHMP approach not only improves prediction performance but also offers an unconventional perspective on motion prediction and allows for a more intuitive understanding of human movement patterns.
Furthermore, our approach can detect regions with more prominent laminar patterns, which are more predictable than those with predominantly turbulent patterns. The extent of laminar dominance within an environment can be used for robot motion planning and exploration tasks.

% explainability of why the prediction is accurate or not, based on location. can be verified with other predictors.

\section{RELATED WORK} \label{section-relatedwork}
The numerous works in trajectory prediction attempt to consider various factors that influence human motion, such as observed dynamics, elements of the static environment, semantic features and social interactions. Based on the underlying principle for the motion model, they can be categorized into pattern-, physics- and planning-based approaches~\cite{rudenko2020human}.

Pattern-based approaches rely on learning patterns and regularities from historical motion data. They use techniques such as Hidden Markov Models, Gaussian Processes and in particular neural networks, to capture temporal dependencies and probabilistic relationships in trajectory data. Recurrent Neural Network (RNN) approaches \cite{alahi2016social,cheng2019human,carrara2019lstm,rudenko2020human} model temporal dynamics and capture non-linear temporal dependencies, while the approaches based on CNNs (Convolutional Neural Networks) extract spatial features that are relevant for motion dynamics~\cite{mohamed2020social,zhao2020noticing, xie2020motion,liu2020trajectorycnn}. Generative models like GANs (Generative Adversarial Network)~\cite{sadeghian2018sophie,fang2022atten}, and CVAEs (Conditional Variational Autoencoder)~\cite{salzmann20,zhou2021sliding,xu2022socialvae} capture dynamic, non-linear dependencies under epistemic uncertainty.
%They provide a measure of uncertainty over predictions and a possibility to draw samples of future motions given the current observed motion trajectory.
Transformer-based architectures introduce attention mechanisms for context understanding~\cite{giuliari2021transformer}. 
%These approaches have shown promise in capturing short to medium-term motion patterns, but they possess inherent limitations, often related to a lack of interpretability or sensitivity to anomalous data.
Many of these approaches primarily focus on predicting stochastic interactions between diverse moving agents in the short-term perspective in scenarios where the effect of the environment topology and semantics is minimal.

%These approaches have shown promise in capturing short to medium-term motion patterns but typically have difficulties handling long-term predictions.

Physics-based approaches construct kinematic models that focus primarily on intrinsic motion dynamics. One such simple yet effective approach is the Constant Velocity Model (CVM), which has demonstrated competitive predictive potential in the short term \cite{scholler2020constant}.
%Notably, Scholler et al.~\cite{scholler2020constant} conducted research revealing that CVM outperforms several state-of-the-art neural predictors when forecasting human motion up to a 4.8-second horizon.
However, CVM falls short when applied to long-term prediction tasks, as it lacks the environmental information, external social interactions and cognitive factors. Other examples of physics-based methods include the Social Force model \cite{helbing1995social,luber2010,farina2017walking}, Reciprocal Velocity Obstacles approaches \cite{van2008reciprocal} and their extensions such as ORCA \cite{van2011reciprocal}, methods based on dynamics such as Switching Linear Dynamical Systems (SLDS) \cite{kooij2018ijcv}. These approaches perform well in certain situations, e.g. for short-term modelling of vehicle dynamics, but, similarly to the CVM, struggle in the long-term perspective.

Planning-based methods, for instance using Markov Decision Processes, have shown a clear potential in the long-term \cite{ziebart2009planning, Rudenko2018icra, rehder2017pedestrian}. Using the map as input, these methods are able to produce long-term non-linear paths towards to distant goals. Still, these methods make optimality assumptions of human motion, which may not always hold in practice. Some further approaches are designed to predict trajectories over extensive durations requiring, however, auxiliary inputs like 
%occupancy grid maps \cite{salzmann20} or 
RGB images \cite{coscia2018long, mangalam2021ynet} to make predictions.

In contrast, our model, LaCE-LHMP, works without explicit knowledge of the environment, and does not require further inputs apart from the observed motion sequence. Instead, it implicitly infers environmental factors and common goals from a representation of observed spatial motion patterns, encoded in the Map of Dynamics (MoD), similarly to CLiFF-LHMP \cite{zhu2023data,zhu2023clifflhmp}. It combines aspects of physics-based and pattern-based approaches, namely the velocity-based transition model and generalization of observed motion. Differently from the prior art, LaCE-LHMP makes the assumption that human motion can be described with laminar-turbulence characteristics similar to airflow and, accordingly, extracts the laminar human motion component to predict motion more accurately.

%%%% Writing about recent long-term human motion prediction method, while they all need map as input
%There are several approaches that focus on estimating goals to achieve long-term prediction. \textcite{ikeda2013modeling} propose to learn sub-goals, specific way-points that pedestrians tend to pass before reaching the final destination and model the long-term behaviour of pedestrians with transition probabilities between sub-goals. The interaction of moving humans with objects in their environment can also be a crucial indicator for predicting their navigation goals. For instance, \textcite{bruckschen2019human} propose to learn a transition model of sequential object interactions and use it to infer the navigation goals using a recursive Bayes' filter. Another example is the work of Mangalam et al.~\cite{mangalam2021ynet}, who present a method to estimate distributions over long-term goals and distributions over a few chosen future waypoints along with the sampled goal points. Their approach requires an RGB scene image as input.

%%%% Will re-write this part, because in related works, some of them are predicitn for 30 s, so this paragraph is not true. But other methods may require extra information like occupancy grid map or RGB image as input, where we make predictions without explicit knowledge about goals and implicitly accounts for the obstacle layout - by using human flow pattern learned from trajs.

\section{METHOD} \label{section-method}
\subsection{Problem statement}

\begin{figure*}[ht]
\centering
\includegraphics[width=1\linewidth]{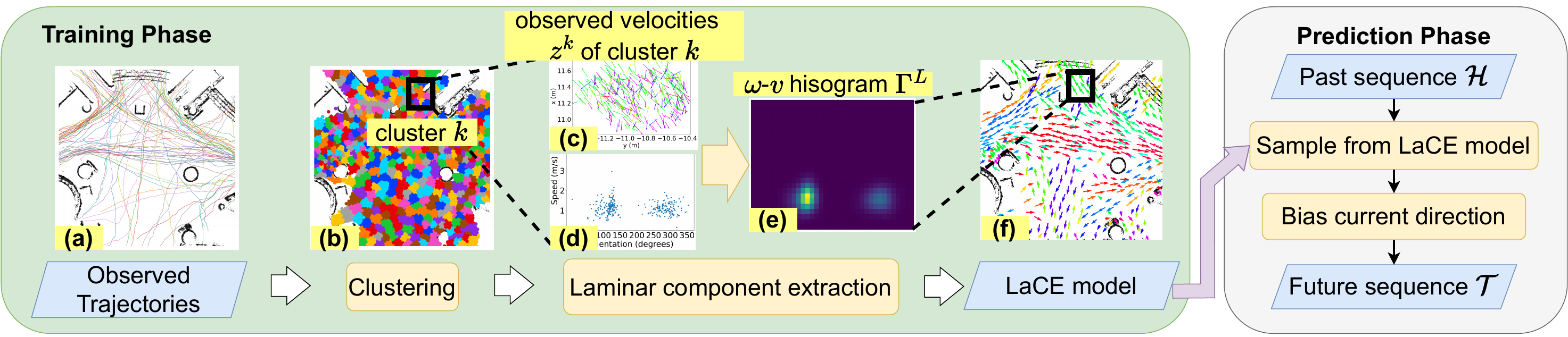}
\caption{Diagram illustrating the training and prediction phases of the LaCE-LHMP approach. In the training phase, observed trajectories \textbf{(a)} are used. Velocity observations, which are depicted in \textbf{(c)} for $(x,y)$ and \textbf{(d)} for $\omega$-$\nu$ distribution, are clustered using K-means into K clusters, shown in \textbf{(b)}. From each cluster's joint $\omega$-$\nu$ distribution, a discrete $\omega$-$\nu$ histogram $\Gamma^R$ is estimated to extract the laminar component $\Gamma^L$, as shown in \textbf{(e)}. The directions with the highest likelihood in $\Gamma^L$ are represented by colored arrows in the LaCE model \textbf{(f)}. The LaCE model is then utilized for prediction.
} \label{fig:master_figure}
\vspace*{-6mm}
\end{figure*}

We frame the task of predicting a person's future trajectory as inferring a sequence of future states. With the input of an observation history of $O_p$ past states of a person, the method predicts $T_p$ future states. The length of the observation history is $O_s \in \mathbb{R}^+$~\SI{}{\second}.
%, equivalent to $O_p \geq 1$ observation time steps.
%
 %(Observations are available and predictions are made with a constant sampling interval of $\Delta t$.)
%
With the current time-step denoted as the integer $t_0 \geq 0$, %$t_0 \in \mathbb{N}$, 
the sequence of observed states is $\mathcal{H} = \langle s_{t_{0} - 1},..., s_{t_{0} - O_p} \rangle$, where $s_t$ is the state of a person at time-step $t$. 
%Referring to a world frame, 
A state is represented by 2D Cartesian coordinates $(x,y)$, direction $\omega$, and speed $\nu$:
%\begin{equation}
$s = (x,y,\omega, \nu)$.
%\end{equation}
From the observed sequence $\mathcal{H}$, we derive the observed speed $\nu_{\mathrm{obs}}$ and direction $\omega_{\mathrm{obs}}$ at time-step $t_0$. Then the current state becomes $s_{t_0} = (x_{t_0},y_{t_0},\omega_{\mathrm{obs}},\nu_{\mathrm{obs}})$.

Given the current state $s_{t_0}$, we estimate a sequence of future states.
Similar to past states, future states are predicted within a time horizon $T_s \in \mathbb{R}^+$~\SI{}{\second}. $T_s$ is equivalent to 
$T_p~\geq~1$ %$T_p \in \mathbb{Z}^+$ 
prediction time steps, assuming a constant time interval $\Delta t$ between two predictions. Thus, the prediction horizon is $T_s = T_p \Delta t$. The predicted sequence is then denoted as $\mathcal{T} = \langle s_{t_0+1}, s_{t_0+2},...,s_{t_0+T_p} \rangle$.

\subsection{Overview of the LaCE-LHMP approach}
The LaCE-LHMP approach \footnote{Code: \url{https://github.com/test-bai-cpu/LaCE-LHMP}} consists of training and prediction phases, as shown in \cref{{fig:master_figure}}. The training phase first extracts the underlying laminar component from the observed trajectories (described in \cref{sec:laminar_extraction}) and learns an MoD, expressed through a set of probabilistic representations of the target area, i.e., the LaCE model. In the prediction phase, both the observed recent trajectory sequence and the learned LaCE model influence the predicted trajectory, depending on the degree of local laminar dominance. In order to select the contributions from both factors depending on the local situation, we propose an adaptive sampling process, see \cref{sec:adaptive_sampling}. Once a likely direction is sampled, the current state can be propagated to predict sequences of future states.

%\begin{figure*}[t]
%\centering
%\includegraphics[width=.60\linewidth]{figures/Untitled Diagram.drawio (4).pdf}
%\caption{Diagram of the training phase and prediction phase of the LaCE-LHMP approach. } \label{fig:master_figure}
%\vspace*{-6mm}
%\end{figure*}

\subsection{Laminar component extraction for enhancing LHMP}
\label{sec:laminar_extraction}
The process of extracting laminar components from observed human trajectories involves three sequential steps:
\begin{enumerate}
    \item \emph{Spatial clustering}: 
    %We use K-means clustering to group velocity observations in the area of interest into $K$ clusters based on their pairwise spatial distances. 
    We apply K-means clustering to group velocity observations within the area of interest into $K$ clusters, by spatial coordinates $(x,y)$ for calculating pairwise distances, as shown in \cref{fig:master_figure}(b). %This method allows us to identify patterns in velocity distribution across different locations. 
    Since the trajectories are not uniformly distributed in the target area, clustering them allows to learn more accurate, representative location-specific flow distributions in both densely and sparsely observed regions.
    
    \item \emph{Local $\omega$-$\nu$ distribution modelling}: Under the assumption that clusters and the respective joint distributions of directions ($\omega$) and speeds ($\nu$) are sufficiently stable over time, 
    %We hypothesize that the joint distribution of directions ($\omega$) and speeds ($\nu$) of the trajectories exhibits temporal relevance among different days within the proximity of the cluster centre position. 
    %In our implementation, 
    we estimate a discrete $\omega$-$\nu$ histogram $\Gamma^R$ to represent each cluster's joint $\omega$-$\nu$ distribution from the observed velocities $z^k_i=(\omega,\nu), i=1,2,3,\dots, N^k$, belonging to cluster $k$, shown in \cref{fig:master_figure}(d). $\Gamma^R$ consists of $N_S$ discrete states, each encapsulating unique combinations of direction and speed. State $J$ represents the estimated probability of a velocity possessing $(\omega_J,\nu_J$).
    %has a total of $N_S$ discrete states, which cover the joint direction and speed combinations. State $J$ then denotes the estimated probability of a velocity possessing $(\omega_J,\nu_J$). 
    Given $N^k$ observations in the cluster, $\Gamma^R$ is given by
    \begin{equation}
      \Gamma^R(J | \mathbf {z}_{1:N^k})=\frac{f_{J, N^k}}{\sum _{j=1}^{N_S} f_{J_j, N^k} } 
     \end{equation}
     where $f_{J_j, N^k}$ is the observed frequency of the $j$-th state at the cluster $k$.
     
    \item \emph{Laminar component extraction}:
    $\Gamma^R$ gives an intuitive sense of the underlying $\omega$-$\nu$ distribution. In reality, $\Gamma^R$ is typically a mixture of more predictable laminar components and ``chaotic'', turbulent components.  Extracting $\Gamma^R$'s laminar component aids unseen trajectory prediction because it is reasonable to assume that its current $(\omega,\nu)$ depends on its recent  $(\omega,\nu)$ as well as the underlying laminar pattern. As shown in \cref{fig:master_figure}(e), $\Gamma ^L(J | \mathbf {z}_{1:t}) $ denotes the laminar component of $\Gamma^R$. $\Gamma ^L(J | \mathbf {z}_{1:t}) $ is estimated using a Bayes filter~\cite{thrun2002probabilistic}, as shown in Alg.~\ref{algo:laminar_component_extraction}, originally from~\cite{bennetts2017probabilistic}.
\begin{algorithm}[t]
\caption{Bayes Filter to extract laminar components}
\label{algo:laminar_component_extraction}
\SetAlgoLined
\KwData{
  Number of states, $N_S$\\
  Total number of observations in the considered cluster $k$, $N^k$\\
  The $i$-th observation, $z_1, z_2, z_{i},\ldots, z_{N^k}$
}
\KwResult{
  Laminar component $\Gamma^L(J | \mathbf{z}_{1:N^k})$ for each $J$ 
}
Initialization:\\
Initialize the prior of ${p}_{J,i}$ without any observation as ${p}_{J,i=0}= 1/{N_S}$ for each $J$\\

\For{$i \leftarrow 1$ \KwTo $N^k$}{
    Calculate $\bar{p}(J, i)$  using the $p(J, i)$  and  $\mathcal {C}(J|J_j,\mathbf {z}_{1:i})$, given by\[
    \bar{p}_{J,i} = \sum_{j=1}^{N_S} {p}_{J_j,i-1} \mathcal{C}(J|J_j,\mathbf{z}_{1:i})\] where $\mathcal {C}(J|J_j,\mathbf {z}_{1:i})$ is a transition model that accounts for the variability of $(\omega,\nu)$ (see Eq.\ref{eq:transition_model});\\
    
    Calculate $p_{J,i}$ using \[
    p_{J,i} = \frac{\bar{p}_{J,t} \mathcal{M}(z_i|J)}{\sum^{N_S}_{j=1}\bar{p}_{J_j,i} \mathcal{M}(z_i|J_j)}\] where $\mathcal{M}(z_i|J_j)$ is a measurement model defined by Eq.~\ref{eq:measurement_model};\\
  
    Calculate $\Gamma^L(J | \mathbf{z}_{1:i})$ using  \[
    \Gamma^L(J | \mathbf{z}_{1:i}) = \frac{\Gamma^L(J | \mathbf{z}_{1:i-1}) + p_{J,i}}{\sum_{j=1}^{N_S} \Gamma^L(J_j | \mathbf{z}_{1:i}) }\] .\\
    }

\KwOut{
  $\Gamma^L(J | \mathbf{z}_{1:N^k})$ for each $J$ given all $N^k$ observations\\
}

\end{algorithm}

\end{enumerate}
The Bayes filter plays a critical role in  Alg.~\ref{algo:laminar_component_extraction} for updating the likelihood for each state $J$. The Bayes filter incorporates the empirical knowledge of the uncertainty with an observed $z=(\omega,\nu)$ using a measurement model given by 
    \begin{equation} \mathcal {M}(z|J)=\frac{1}{2 \pi \sigma _\omega \sigma _\nu }\mathbf {exp}-\Bigg (\frac{\Delta (\omega,J_\omega)^2}{2\sigma _\omega ^2} + \frac{|\nu -J_\nu |^2}{2\sigma _\nu ^2}\Bigg) 
    \label{eq:measurement_model}
    \end{equation}
where $\Delta(\omega, J_\omega)$ and $|\nu -J_\nu |$ correspond to great-circle distance and spatial distance (e.g., Euclidean distance) between measurement $z=(\omega,\nu)$ and the state $J(J_\omega, J_\nu)$, respectively. Parameters $\sigma_\omega$ and $\sigma_\nu$  correspond to the confidence intervals with respect to the variables of direction variable and speed, which can be set empirically. The measurement model is used in the transition model given by Eq.~\ref{eq:transition_model}:
    \begin{equation} 
    \mathcal {C}(J|J_j,\mathbf {z}_{1:i})=\mathcal {C}(J|J_j,z_{1:i-1})+\mathcal M(z_i|J)
    \label{eq:transition_model}
    \end{equation}
which assigns a posterior to each state $J$ based on the frequency of the transition between $J$ and another state $J_j$. This transition model enables the suppression of the posteriors of states associated with intermittent transitions, thereby enhancing the discernibility of laminar dominant states. A comparison of the visualized raw trajectory data, the corresponding probabilistic representation $\Gamma^R$, and the extracted laminar component $\Gamma^L$ is provided in Fig.~\ref{algo:laminar_component_extraction}.

\subsection{Adaptive sampling based on Laminar-dominant condition}
\label{sec:adaptive_sampling}
Considering that both the observed part of the trajectory and the underlying laminar pattern can contribute to the prediction, the our method involves a trade-off between relying on the laminar component or adapting to recent observations. For this reason, it is useful to quantify the degree of laminar dominance to guide the trade-off. In our proposed approach, the Kullback-Leibler (KL) divergence between $\Gamma^R$ and $\Gamma^L$, denoted as $D_{KL}(\Gamma^R \parallel \Gamma^L)$, serves as an indicator of thelaminar dominance. A larger divergence value corresponds to a lower degree of laminar dominance. 

In the prediction phase, to estimate $\mathcal{T}$, for each prediction time step, we sample a direction from the laminar component corresponding to the current position. Assuming that a person tends to continue walking at the same speed as in the last time step, we bias the direction of motion with the direction $\omega_s$ sampled from $\Gamma^L$, as $\omega_t = \omega_{t-1} + (\omega_s - \omega_{t-1}) \cdot K(\omega_{s} - \omega_{t-1})$, where $K(\cdot)$ is a kernel function that defines the degree of impact of the sampled direction. To define the degree of laminar dominance, we employ a Gaussian kernel with the KL divergence serving as the kernel width, $K(x) =  e ^ {-\beta \left\Vert x \right\Vert ^ 2}$, where $\beta = 10^{D_{KL}(\Gamma^R \parallel \Gamma^L)}$. When there is high divergence, indicating low laminar dominance at the current location, the proposed method tends to behave more like a CVM. Conversely, with smaller divergence, suggesting the position is likely to be laminar dominated, and therefore, the prediction will align more with a laminar pattern.

\section{EXPERIMENTS} \label{section-experiments}
This section describes the experimental setup for qualitative and quantitative evaluation of the proposed approach. 

\textbf{Dataset}: For evaluation, we use the ATC shopping mall dataset \cite{brvsvcic2013person}. This dataset covers a large indoor environment with a total area of around \SI{900}{\metre\squared}.
%The dataset recorded human trajectories for 92 days.
Given the immense length of the ATC dataset (92 days), we use a subset of 10 days in the experiments, with the first day for training and the remaining 9 days for evaluation. For both our methods and baselines, we use the same training and evaluation data.

\textbf{Baselines}: We compare the performance of our approach with three baselines: CLiFF-LHMP, Trajectron++ and CVM. 

%CLiFF-LHMP approach is a method for predicting long-term human motion trajectories using maps of dynamics (MoDs) that encode spatial motion patterns in the environment. It uses a multi-modal probabilistic representation of a velocity field (CLiFF-map), which is built from sparse and incomplete observations of human motion. In the prediction process, the CLiFF approach samples directions from the CLiFF-map at each predicted location and biases a constant velocity prediction with the sampled direction, using a Gaussian kernel function to balance the influence of the two components. 
CLiFF-LHMP \cite{zhu2023clifflhmp}, similarly to LaCE-LHMP, is based on Maps of Dynamics, but uses a different representation of human motion, namely the CLiFF-map. Differently from LaCE-LHMP, the CLiFF-map has a regular grid structure and uses Gaussian Mixture Modelling to detect the dominant motion patterns in each grid cell. Importantly, it does not factor our the turbulent component of the training data. Previously, the CLiFF approach was validated with the ATC dataset~\cite{brvsvcic2013person} compared to a vanilla LSTM model as a baseline~\cite{zhu2023data}. The results show that the CLiFF approach outperforms the LSTM model at the long prediction horizons of up to 60 seconds in terms of the average and final displacement errors. The superior performance of CLiFF-LHMP makes it a suitable baseline algorithm for our comparative study.

Trajectron++ (T++) \cite{salzmann20} represents a state-of-the-art approach employing a graph-structured generative neural network based on a conditional-variational autoencoder. To run T++ we used public code and trained the model for 100 epochs on the training day of ATC dataset. Parameter configurations are provided with project code.

\begin{figure}[h]
\centering
\includegraphics[width=.49\linewidth]{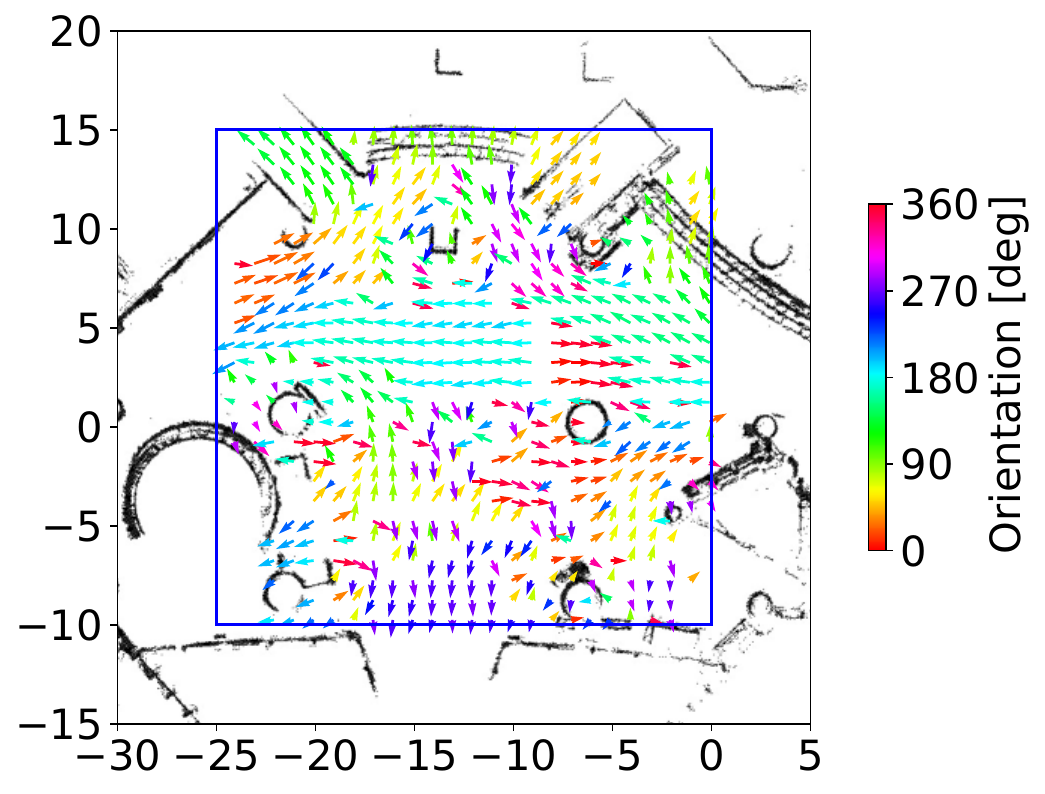}
\includegraphics[width=.49\linewidth]{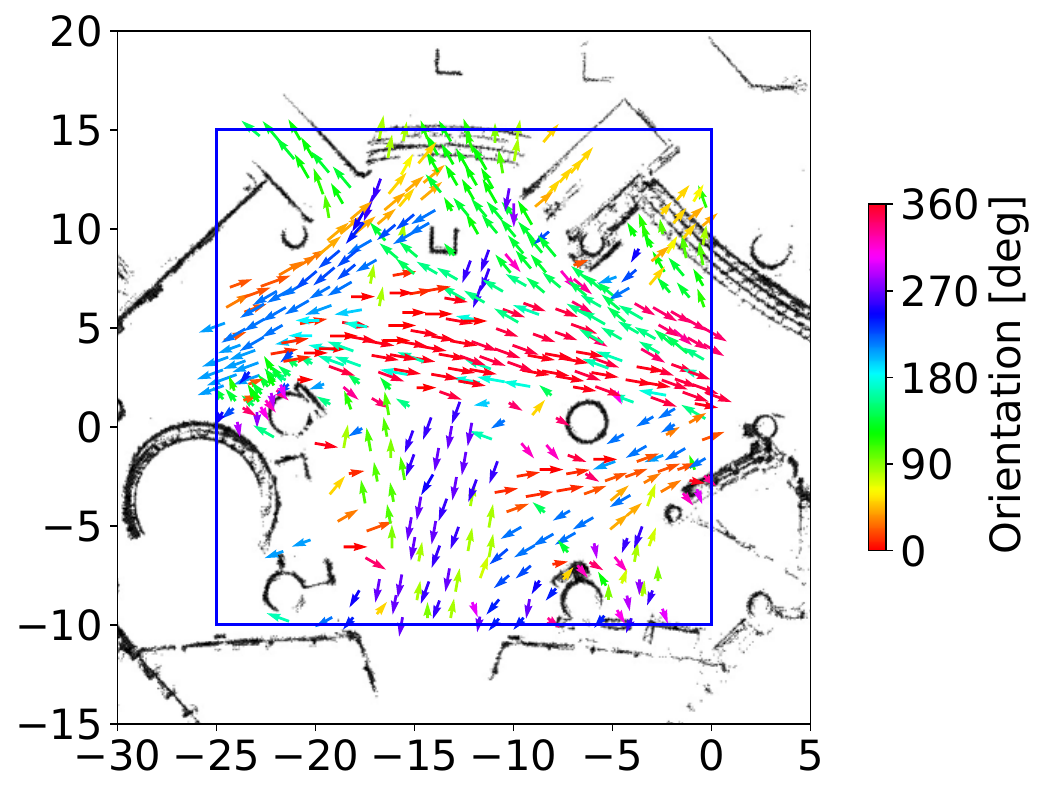}
\caption{CLiFF-map (\textbf{left}) and LaCE model (\textbf{right}) are shown in colored arrows. In the CLiFF-map, arrows show the mean value of the component with the highest weights. In the LaCE model, arrows show the directions with the highest likelihood.}
\label{fig:mod_compare}
\vspace*{-4mm}
\end{figure}

\textbf{Implementation details}: Given the map of the ATC environment, we focus on the central square area highlighted by the blue square in \cref{fig:mod_compare}. This area has dimensions where $X$ ranges from $-$25 to 0 and $Y$ ranges from $-$10 to 15, amounting to an area of \SI{625}{\metre\squared}. In contrast to the east corridor, the central square offers a more open space, allowing pedestrians greater freedom of movement and presenting more obstacles. Conversely, the human flow patterns in the east corridor are simpler and more restricted.

In the experiments, the ATC dataset was downsampled to 1 Hz. We use an observation horizon of \SI{3}{\second} for input and the following trajectory (up to \SI{20}{\second} long) as the ground truth.

%%%%%% Parameter selection %%%%%%%%%
For parameter settings of all the methods, prediction horizon $T_s$ is set to 1--20 \si{\second}, and observation horizon $O_s$ is \SI{3}{\second}. Prediction time step $\Delta t$ is set to \SI{1}{\second}. In the experiment, the values of $\omega_{\mathrm{obs}}$ and $\nu_{\mathrm{obs}}$ are calculated as a weighted sum of the finite differences in the observed state, as in the recent ATLAS benchmark \cite{rudenko2022atlas}. 
With the same parameters as in~\cite{rudenko2022atlas}, the sequence of observed velocities is weighted with a zero-mean Gaussian kernel with $\sigma = 1.5$ to put more weight on more recent observations, such that $\omega_{\mathrm{obs}} = \sum_{t=1}^{O_p}\omega_{t_0 - t}g(t)$ and $\nu_{\mathrm{obs}} = \sum_{t=1}^{O_p}v_{t_0 - t}g(t)$, where $g(t) = %\frac{1}{\sigma\sqrt{2\pi}e^{-\frac{t^2}{\sigma}}}$.
%(\sigma\sqrt{2\pi}e^{-\frac{t^2}{\sigma}})^{-1}$.
(\sigma\sqrt{2\pi}e^{\frac{1}{2}(\frac{t}{\sigma})^2})^{-1}$. For the LaCE-LHMP experiment, in the spatial clustering step, cluster number $K$ is set to 500, targeting each cluster region to cover approximately \SI{1}{\metre\squared}
. In constructing the discrete $\omega$-$\nu$ histogram, speed ($\nu$) bins are defined at \SI{0.2}{\metre\per\second} intervals, ranging from 0 to \SI{5}{\metre\per\second}, and direction ($\omega$) bins are defined at 10-degree intervals, covering the full 360-degree range.

\textbf{Evaluation metrics}: For the evaluation of the predictive performance, we used the following metrics: \emph{Average} and \emph{Final Displacement Errors} (ADE and FDE) and \emph{Top-k ADE/FDE}. ADE describes the mean $L^2$ distance between predicted trajectories and the ground truth. FDE describes the $L^2$ distance between the predicted final position and the ground truth final position at the last prediction time step. \emph{Top-k ADE/FDE} compute the displacements between the ground truth position and the closest of the $k$ predicted trajectories. $k$ is set to 5 in the evaluation.

For evaluating T++, we use the most-likely output configuration, which generates deterministic and most-likely single output.
%For presenting ADE/FDE results, we use the most likely output trajectory of LaCE-LHMP, CLiFF-LHMP, and T++. 
When evaluating our approach, for any given observed sequence, LaCE-LHMP can be executed multiple times to randomly generate a set of predicted trajectories. Based on practical applications for autonomous robots, LaCE-LHMP can rank these predicted trajectories and provide the most likely output. The probability of the output sequence $\mathcal{T}$ is calculated as the product of probabilities of samples taken from histograms over $\mathcal{T}_p$ prediction time steps. A higher probability results in a higher ranking. For CLiFF-LHMP, we determine the likelihood of the sampled velocity using the probability density function of the Semi-Wrapped Gaussian Mixture Model distribution. %In the experiments, both approaches were executed 5 times and output the highest ranking trajectory.
For robustness evaluation, both CLiFF-LHMP and LaCE-LHMP are run 10 times and standard deviations are shown in \cref{tab:expres}.

%For evaluating the top-k ADE/FDE values for our approach, CLiFF-LHMP and T++, that exhibit multimodality, we set $k=5$, thus generating 5 predictions for each ground truth trajectory.

\section{RESULTS} \label{section-results}
%In this section, we present the results obtained in the ATC dataset with our approach compared to three baselines. The performance is evaluated with both quantitative and qualitative analysis.

\begin{table}[t]
    \centering
    \begin{tabular}{lll}
     \toprule
        \textbf{Method} & \textbf{ADE / FDE} & \textbf{\begin{tabular}[c]{@{}c@{}}Top-k \\ ADE / FDE\end{tabular}} \\
    \midrule
        CVM & 4.26 / 9.01 & -  \\
        Trajectron++ & 6.09 / 12.86 & \textbf{2.96} / \textbf{5.86}  \\
        CLiFF-LHMP & 3.52$\pm$0.009 / 7.40$\pm$0.021 & 3.00 / 6.09  \\
        LaCE-LHMP (Ours) & \textbf{3.31$\pm$0.006} / \textbf{6.93$\pm$0.013} & 3.00 / 6.13  \\
        \bottomrule
    \end{tabular}
    \caption{Long-term prediction (\SI{20}{\second}) results on the ATC dataset. With $O_s = \SI{3}{\second}$, errors are reported as ADE/FDE in meters.}
    \label{tab:expres}
\vspace*{-5mm}
\end{table}

\begin{figure}[h]
\centering
\includegraphics[width=.47\linewidth]{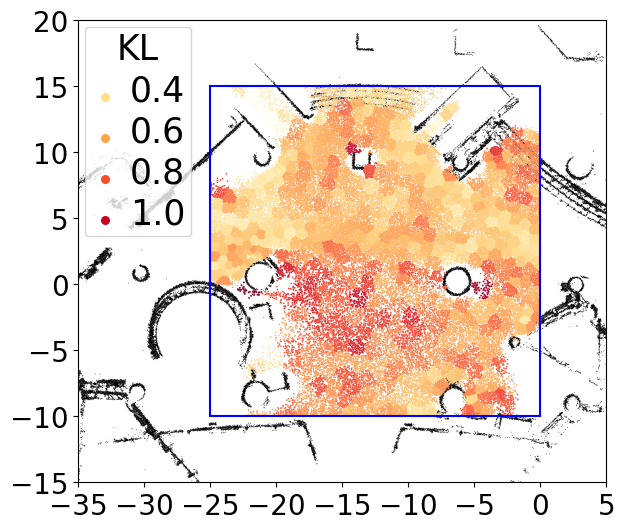}
\includegraphics[width=.505\linewidth]{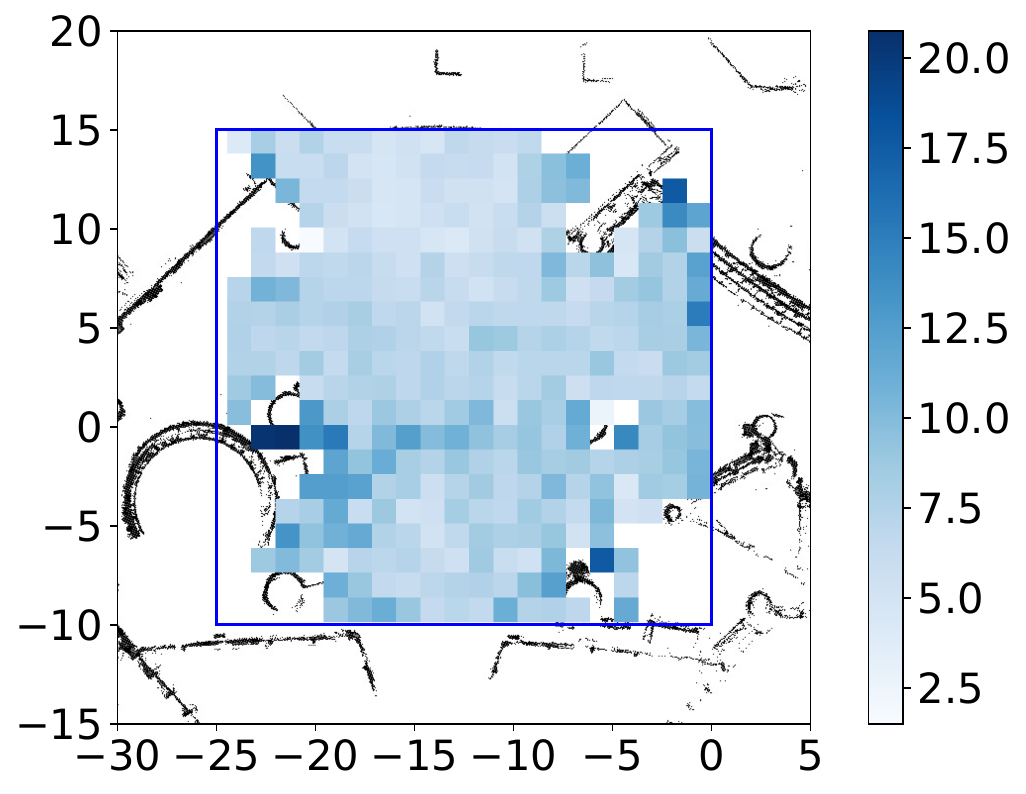}
\caption{\textbf{Left:} KL divergence between $\Gamma^R$ and $\Gamma^L$.  \textbf{Right:} A heatmap illustrating the FDE values of LaCE-LHMP in the ATC dataset, with a prediction horizon of \SI{20}{\second}. Predictions exhibit higher accuracy in the central region. Predictions exhibit higher accuracy in the central region, which is predominantly laminar, as indicated by lower KL divergence.}
\label{fig:FDE_heatmap_bin}
\vspace*{-4mm}
\end{figure}

\subsection{Quantitative results}

\begin{figure}[ht]
\centering

\includegraphics[width=.49\linewidth]{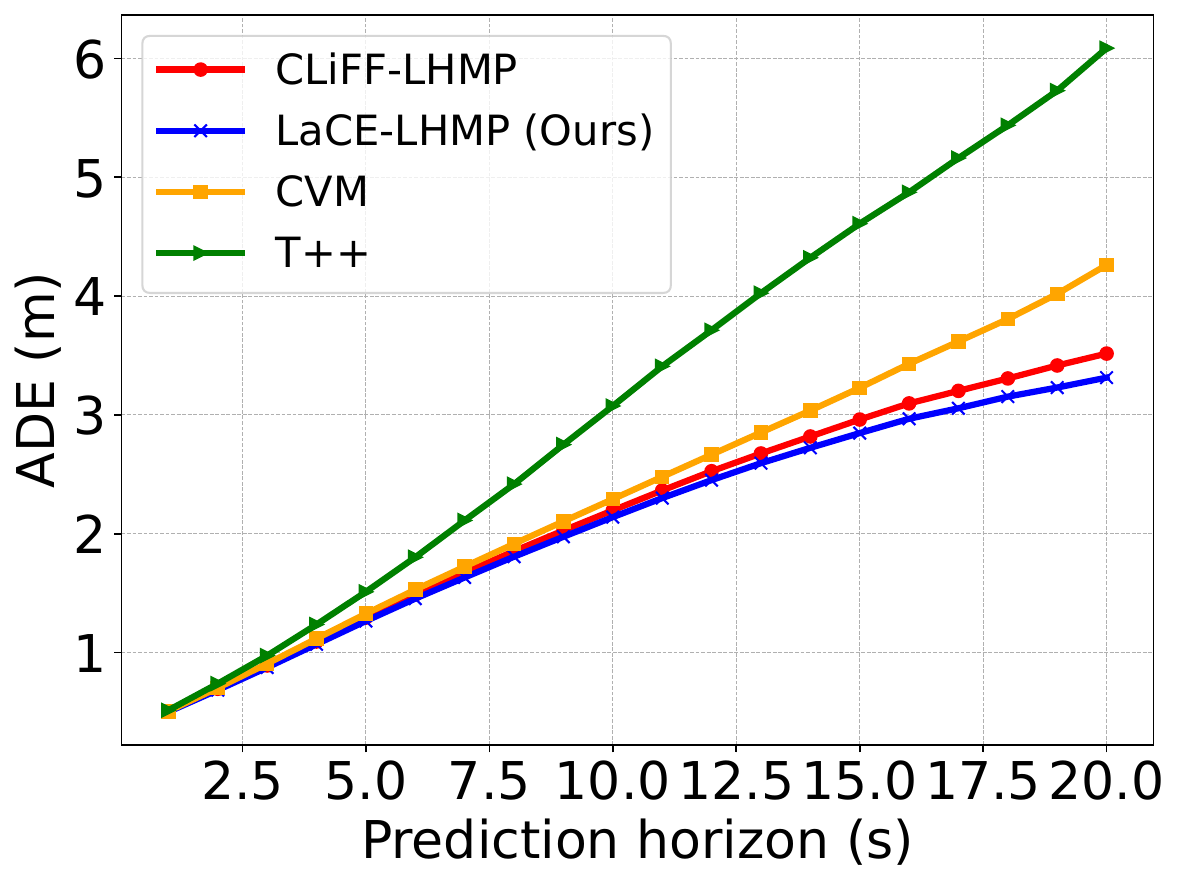}
\includegraphics[width=.49\linewidth]{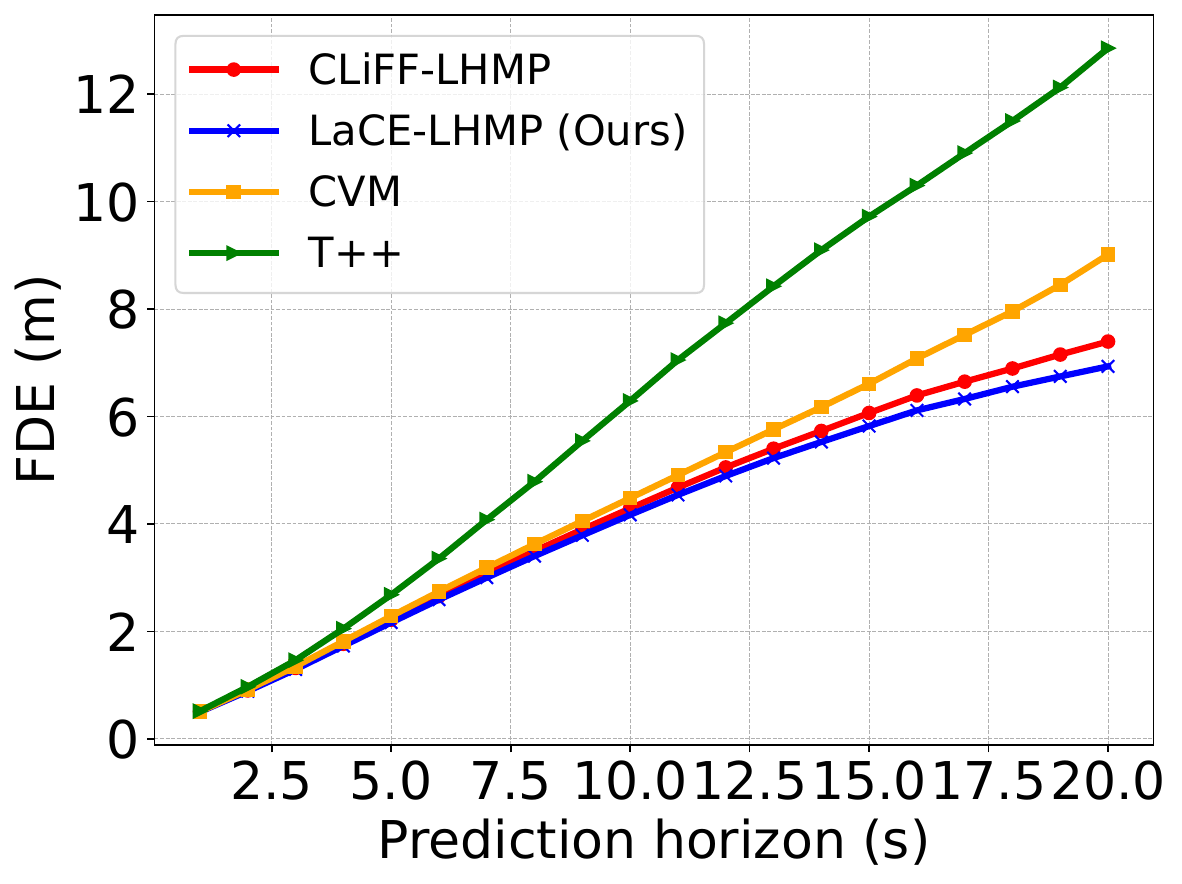}
\includegraphics[width=.49\linewidth]{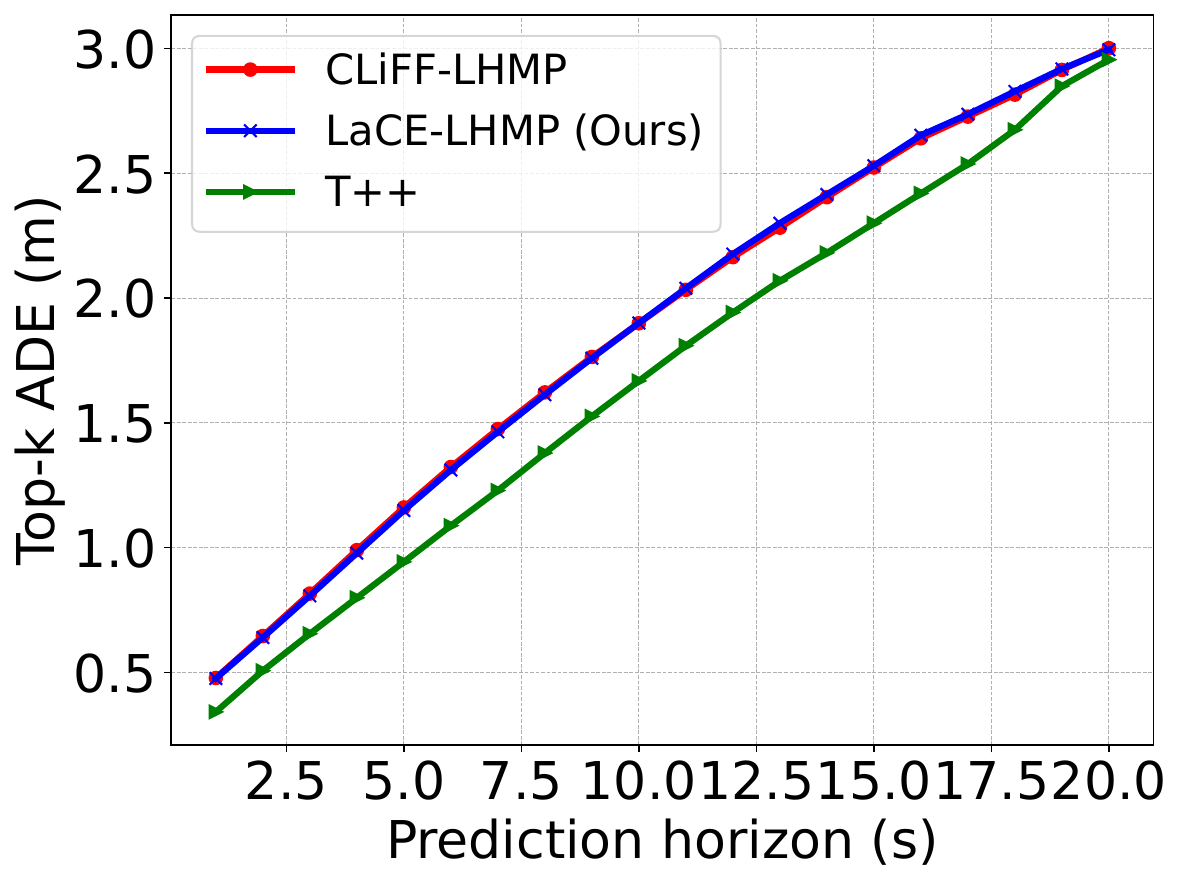}
\includegraphics[width=.49\linewidth]{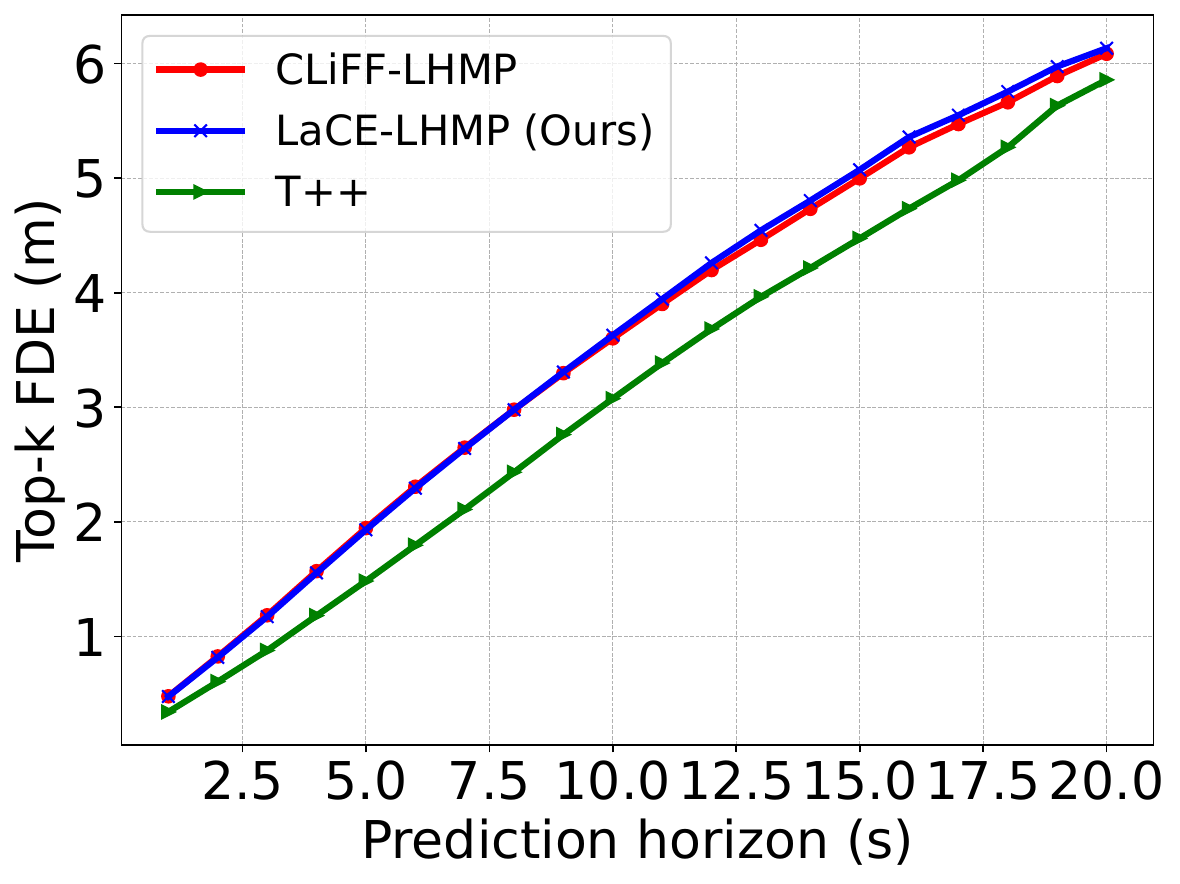}

\caption{ADE/FDE (\textbf{top}) and top-k ADE/FDE (\textbf{bottom}) in the ATC dataset with a prediction horizon 1--\SI{20}{\second}. Predictions with the LaCE model are more accurate during the whole considered period, as indicated by lower ADE/FDE values, which signify improved performance.}
\label{fig:adeandfde}
\vspace*{-4mm}
\end{figure}

We compare LaCE-LHMP with CLiFF-LHMP, T++, and CVM with prediction horizon from \SI{1}{\second} to \SI{20}{\second}. \cref{fig:adeandfde} shows the quantitative results obtained in the ATC dataset described above.  ADE/FDE and top-k ADE/FDE values for predictions using the LaCE-LHMP and baselines are presented. In the short-term perspective, all approaches perform on par. As the prediction horizon increases, LaCE-LHMP increasingly improves in terms of accuracy over baseline approaches. Notably, our method achieves significantly higher accuracy in the considered period. For the minimum ADE and FDE value from 5 randomly sampled trajectories, T++ has achieved better performance of top-k ADE/FDE, but in the long-term prediction horizon of \SI{20}{\second}, our approach performs on par. With effective ranking of predicted trajectories, our method outperforms baselines in ADE/FDE values.

%In the experiments, we ran both CLiFF-LHMP and our approach 10 times. For CLiFF-LHMP, the standard deviations of ADE and FDE, respectively, are 0.009 and 0.021 at a prediction horizon of \SI{20}{\second}. For LaCE-LHMP, the standard deviations of ADE and FDE are lower: 0.006 and 0.013.

\cref{tab:expres} summarises the performance results of our method against the baseline approaches at the maximum prediction horizon of \SI{20}{\second}. At \SI{20}{\second} in the ATC dataset, our method achieves a 6.0\% ADE and 6.4\% FDE improvement in performance compared to CLiFF-LHMP, and 45.6\% ADE and 46.1\% FDE compared with Trajectron++. At the same time, LaCE-LHMP achieves a comparable top-k ADE and a slightly larger top-k FDE value compared with T++.

To evaluate the relation between prediction performance and the degree of laminar dominance in the environment, we present a heatmap of FDE values of our approach for prediction horizon \SI{20}{\second} in \cref{fig:FDE_heatmap_bin}. In laminar-dominated regions, predictions made using the LaCE model are more accurate than in regions with more turbulent patterns, indicating that the former are more predictable.

%\begin{figure}
%\centering
%\includegraphics[width=.49\linewidth]%{figures/res/ADE.pdf}
%\includegraphics[width=.49\linewidth]{figures/res/FDE.pdf}
%\caption{ADE/FDE in the ATC dataset with a prediction horizon 1--\SI{20}{\second}. Predictions with the LaCE model are more accurate during the whole considered period.}
%\label{fig:adeandfde}
%\vspace*{-1mm}
%\end{figure}

\begin{figure}
\centering

\includegraphics[width=.49\linewidth]{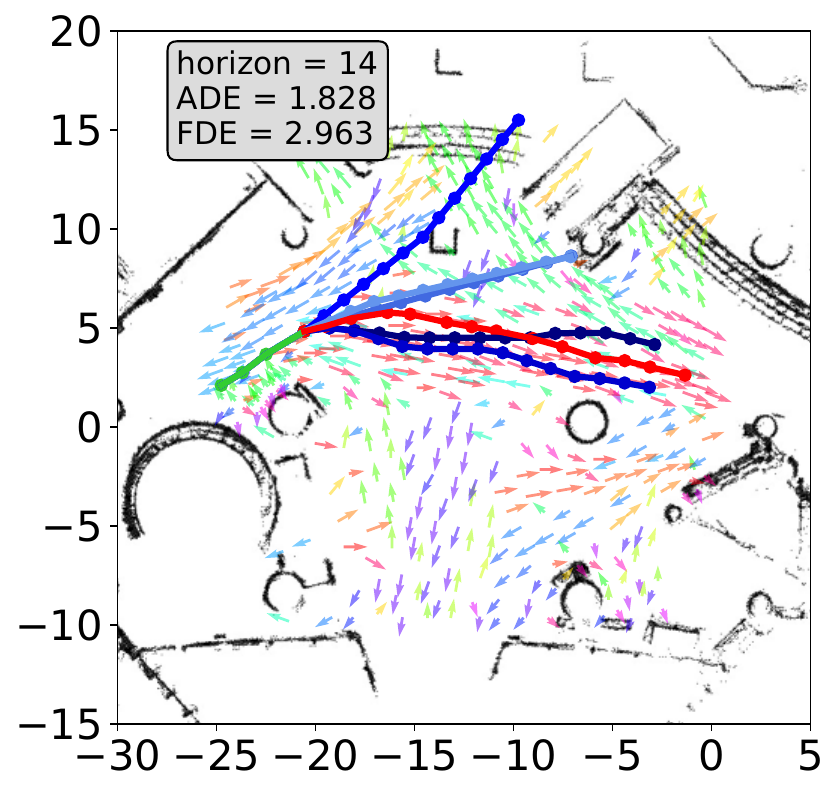}
\includegraphics[width=.49\linewidth]{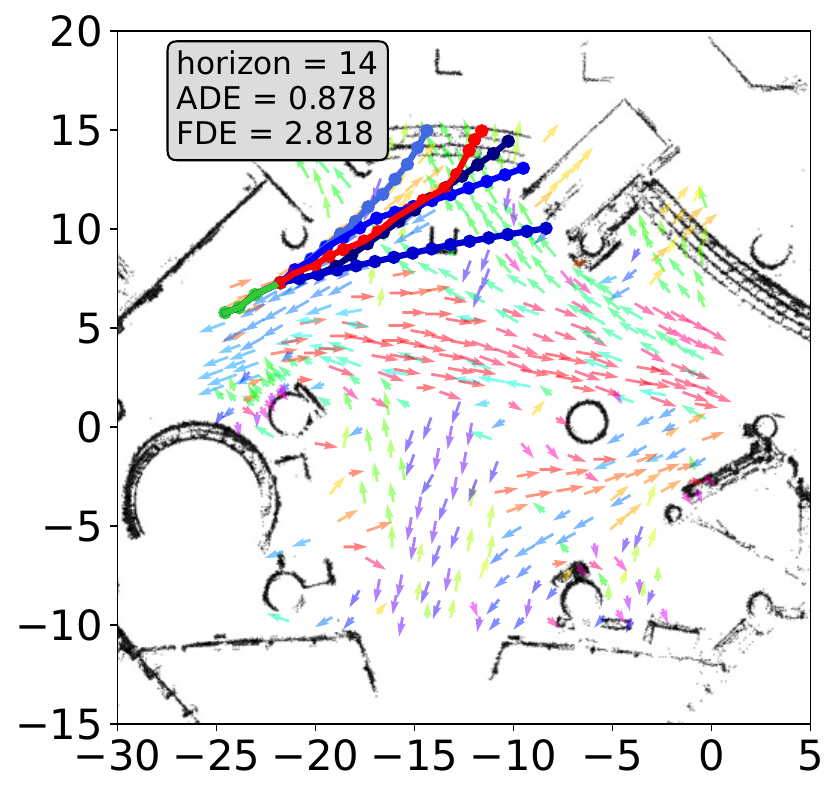}

\caption{Two examples of predicted trajectory rankings using LaCE-LHMP. The \textbf{red} line represents the ground truth trajectory, the \textbf{green} line represents the observed trajectory and \textbf{blue} lines the predicted trajectories, with darker shades of blue indicating higher-ranked predictions. The LaCE model is shown in colored arrows. Predictions that align more closely with dominant flow patterns, and are thus darker in blue, demonstrate higher accuracy, showcasing the effectiveness of the ranking mechanism.}
\label{fig:rank}
\vspace*{-4mm}
\end{figure}

\begin{figure}
\centering
\begin{tikzpicture}
  % Upper left image
  \node[anchor=south west,inner sep=0] (image1) at (0,0) {\includegraphics[clip,trim=0mm 9.5mm 0mm 0mm,height=35mm]{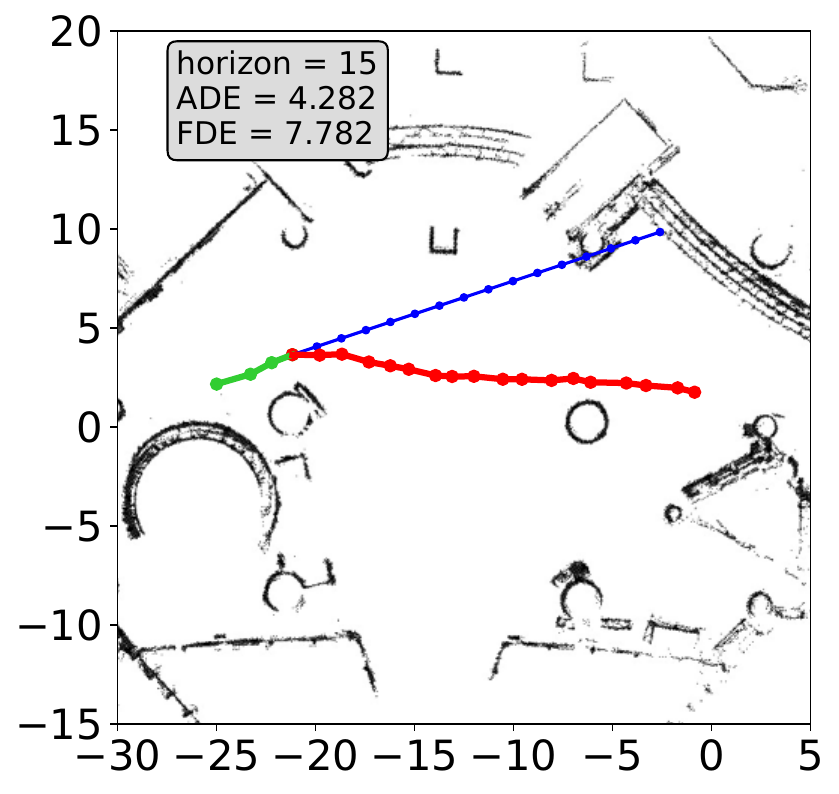}};
  \node[font=\bfseries\footnotesize, text=darkgray, fill=yellow!30] at  (image1.center) [xshift=14mm, yshift=14.5mm] {CVM}; % Position text over image
  
  % Upper right image
  \node[anchor=south west,inner sep=0] (image2) at (0.5\linewidth,0) {\includegraphics[clip,trim=20mm 9.5mm 0mm 0mm,height=35mm]{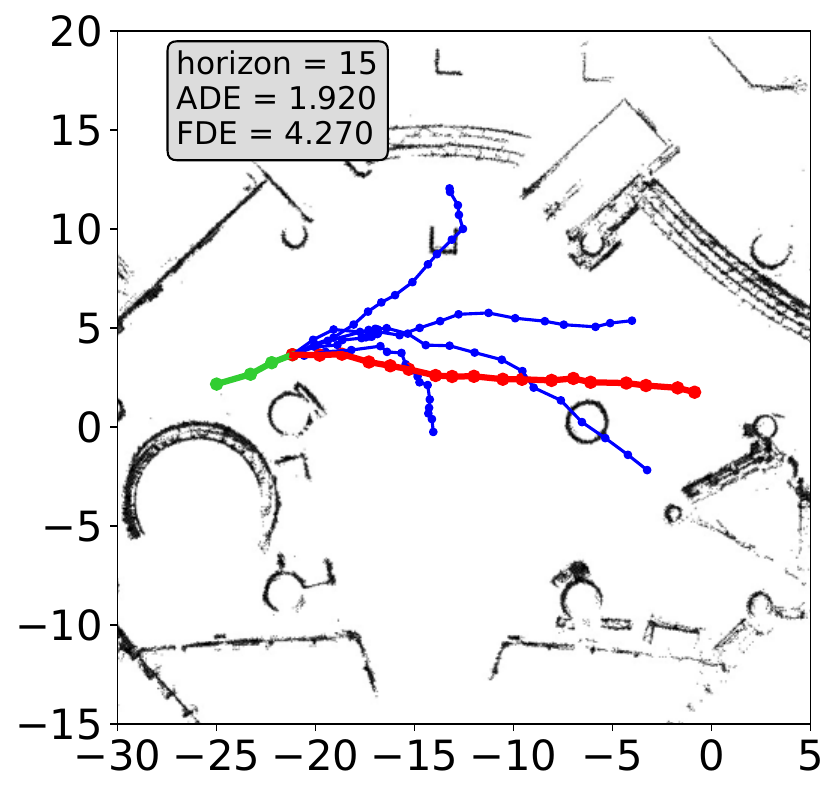}};
  \node[font=\bfseries\footnotesize, text=darkgray, fill=yellow!30] at (image2.center) [xshift=12mm, yshift=14.5mm] {T++}; % Position text over image
  
  % Lower left image
  \node[anchor=south west,inner sep=0] (image3) at (0,-\linewidth*0.40*1.1) {\includegraphics[clip,trim=0mm 0mm 0mm 0mm,height=37mm]{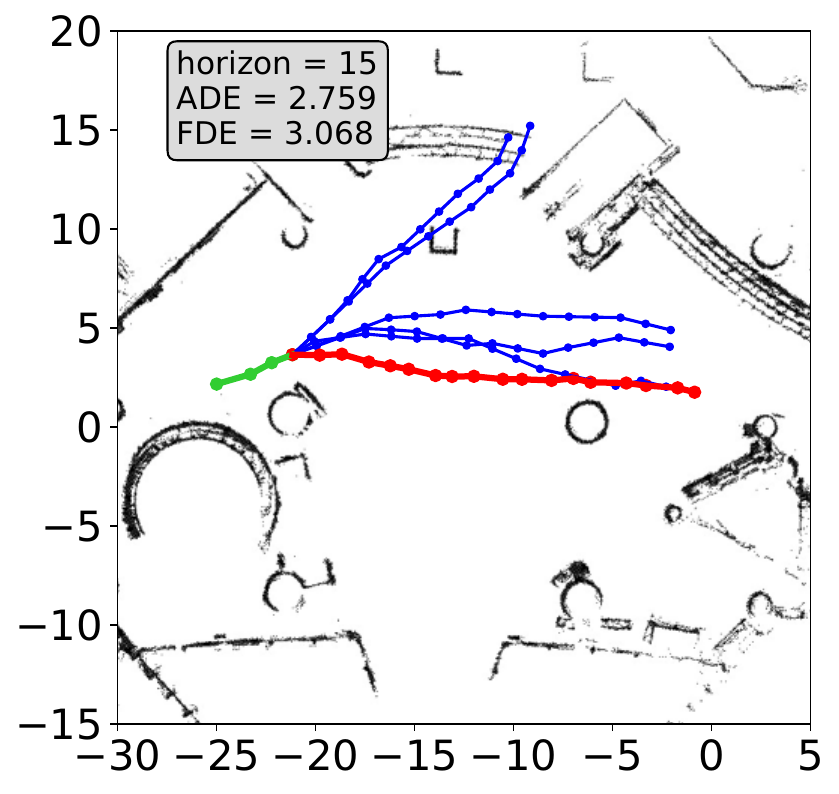}}; % Adjusted for spacing
  \node[font=\bfseries\footnotesize, text=darkgray, fill=yellow!30] at (image3.center) [xshift=8.5mm, yshift=16.5mm] {CLiFF-LHMP}; % Position text over image
  
  % Lower right image
  \node[anchor=south west,inner sep=0] (image4) at (0.5\linewidth,-\linewidth*0.40*1.1) {\includegraphics[clip,trim=20mm 0mm 0mm 0mm,height=37mm]{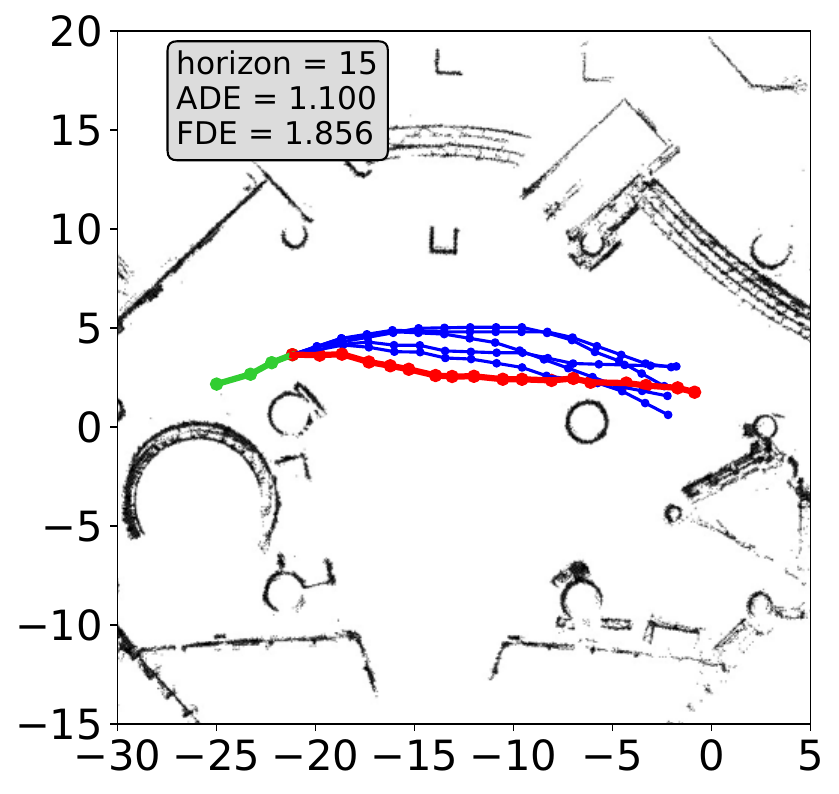}}; % Adjusted for spacing
  \node[font=\bfseries\footnotesize, text=darkgray, fill=yellow!30] at (image4.center) [xshift=6mm, yshift=16.5mm] {LaCE-LHMP}; % Position text over image
\end{tikzpicture}
\caption{Prediction examples in ATC with \SI{15}{\second} prediction horizon. \textbf{Red} line represents the ground truth trajectory, \textbf{green} line represents the observed trajectory, and the \textbf{blue} lines represent the predicted trajectories. \textbf{Upper-left}: Prediction result of CVM, with the highest prediction error in this case. \textbf{Upper-right}: Prediction result of T++. As no explicit map is provided, the predicted trajectory might collide with obstacles. \textbf{Lower-left}: Prediction result of CLiFF-LHMP. Despite not using an obstacle map, the predicted trajectories avoid obstacles. CLiFF-LHMP captures multiple flow patterns, such as passing through stairs (top of the map) and moving to the right, resulting in a broader spread of predictions. \textbf{Lower-right}: Prediction result of LaCE-LHMP. Compared to the baselines, our approach better captures the dominant flow patterns. The generated trajectories are more concentrated, closely following the dominant flow and yielding a more accurate prediction.}

\label{fig:example1}
\vspace*{-6mm}
\end{figure}
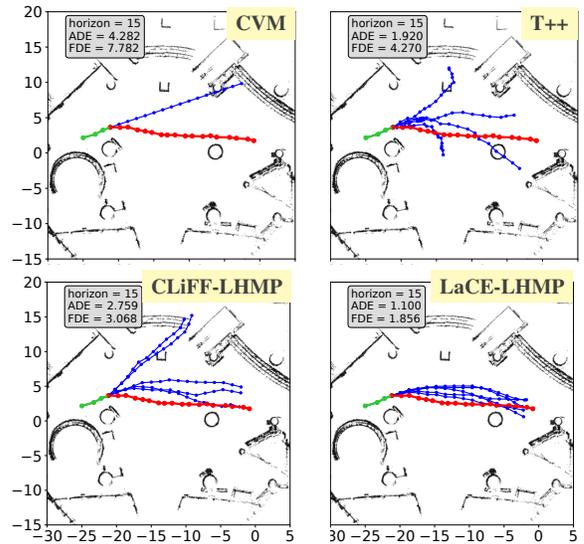

\subsection{Qualitative results}

We present the CLiFF-map and LaCE model depicting human flow patterns within the central area of the ATC dataset in \cref{fig:mod_compare}. The CLiFF-map describes the human flow at a given location with a multimodal distribution, while the LaCE model reveals the dominant human flow patterns, achieved by estimating $\Gamma^L$ as a probabilistic representation of the Laminar component for each cluster. The difference between the two methods can be found in the bottom area in the middle of both subfigures in \cref{fig:mod_compare}. Both the LaCE model and CLiFF-map present a horizontal human flow in the middle of the scene. While in the bottom area, from the LaCE model, one can observe a clear motion pattern originating from the top and progressing toward the bottom. In contrast, the corresponding area in the CLiFF-map exhibits a less distinct flow pattern. 

\cref{fig:rank} demonstrates ranking predicted trajectories using LaCE-LHMP. The predictions with a higher ranking (in a darker blue colour) align with the dominant flow pattern in the LaCE model and the highest-ranked prediction is closer to the ground truth.
We present examples of predictions in \cref{fig:example1}. With CLiFF-map, the predictions exhibit a spread due to the multimodal distribution within the map. In contrast, predictions made with the LaCE model result in more concentrated trajectories, aligning with the dominant flow patterns observed in the middle area of the LaCE model.

%\begin{figure}
%\centering
%\includegraphics[width=.98\linewidth]{figures/examples/cliff_9453700.png}
%\includegraphics[width=.98\linewidth]{figures/examples/LT_9453700.png}
%\caption{A prediction example in ATC with \SI{16}{\second} prediction horizon. When the ground truth trajectory does not follow the dominant flow patterns, with LE map, the predictions have higher displacement error. With CLiFF-map, generated trajectories can capture this specific case and reach a more accurate prediction.}
%\label{fig:example2}
%\vspace*{-5mm}
%\end{figure}

%\cref{fig:example2} presents another scenario where the ground truth trajectory deviates from the dominant flow patterns. In this case, predictions using the LE map tend to exhibit higher displacement errors. However, the CLiFF-map, with its capacity to capture various flow patterns, offers a more accurate prediction in such specific cases.

%Two examples are shown in \cref{fig:example1} and \cref{fig:example2}. In \cref{fig:example1}, when predicting with CLiFF-map, due to multimodal distribution in CLiFF-map, the generated trajectories follow multiple flow patterns and become more spread. When predicting with LE map, the generated trajectories are more concentrated because of the dominant flow patterns in the middle area in the LE map. In \cref{fig:example2}, when the ground truth trajectory does not follow the dominant flow patterns, with LE map, the predictions have higher displacement error. With CLiFF-map, generated trajectories can capture this specific case and reach a more accurate prediction.

\section{CONCLUSION} \label{section-conclusions}
In this study, we introduce a novel approach inspired by airflow modelling to tackle the challenging problem of long-term human motion prediction (LHMP). Our proposed Laminar Component Enhanced (LaCE) LHMP approach is designed to extract the laminar component of human trajectories, creating a probabilistic representation of the underlying streamlined and predictable flows. This approach improves the prediction of future motion patterns substantially. In addition to the laminar flow component extraction, another key innovation of the LaCE approach is its utilization of KL-divergence to quantitatively measure the laminar-dominant condition, allowing for adaptive adjustments to the contribution of the laminar component in the prediction process. The degree of laminar dominance can indicate the level of predictability of human motion in the environment.

The promising results in a benchmark against the prior art LHMP methods 1) confirm that laminar flow is a useful category to analyze real-world human trajectories; 2) support our hypothesis that laminar flow components are distinguishable in human motion patterns; 3) demonstrate the superior prediction performance of the LaCE-LHMP approach; and 4) show that laminar-dominant measurement can quantitatively indicate the predictability of the regions, contribute to a better understanding of human movement patterns.

In the future work, we intend to study more closely the turbulent component of the model, which can be used to describe, detect and predict abnormal behavior.

\printbibliography
\end{document}